\documentclass[11 pt,a4paper]{article}
\usepackage[utf8]{inputenc}
\usepackage{amsmath}
\usepackage{amsfonts}
\usepackage{amssymb}
\usepackage{pgfplots}
\usepackage{tikz}
\usepackage{subcaption}
\usepackage{amsmath}
\usepgfplotslibrary{fillbetween}
\usetikzlibrary{patterns}
\usepackage{indentfirst}
\usepackage{amsmath}
\usepackage{booktabs}
\usepackage{xcolor}
\usepackage{pgfplotstable}
\usepackage{lscape}
\usepackage{mathtools}
\usepackage[paper=portrait,pagesize]{typearea}

\author{Dr. Jacques Balayla MD, MPH, CIP, FRCSC\footnote{To whom correspondence should be addressed: Dr. Jacques Balayla MD, MPH, CIP, FRCSC. Quilligan Scholar. e-mail: jacques.balayla@mcgill.ca. Department of Obstetrics and Gynaecology. McGill University, Montreal, Quebec, Canada}}

\title{\LARGE Prevalence Threshold and bounds in the Accuracy of Binary Classification Systems}

\date{}

\begin{document}
\maketitle  
\begin{abstract}
The accuracy of binary classification systems is defined as the proportion of correct predictions - both positive and negative - made by a classification model or computational algorithm. A value between 0 (no accuracy) and 1 (perfect accuracy), the accuracy of a classification model is dependent on several factors, notably: the classification rule or algorithm used, the intrinsic characteristics of the tool used to do the classification, and the relative frequency of the elements being classified. Several accuracy metrics exist, each with its own advantages in different classification scenarios. In this manuscript, we show that relative to a perfect accuracy of 1, the positive prevalence threshold ($\phi_e$), a critical point of maximum curvature in the precision-prevalence curve, bounds the $F{_{\beta}}$ score between 1 and 1.8/1.5/1.2 for $\beta$ values of 0.5/1.0/2.0, respectively; the $F_1$ score between 1 and 1.5, and the Fowlkes-Mallows Index (FM) between 1 and $\sqrt{2} \approx 1.414$. We likewise describe a novel $negative$ prevalence threshold ($\phi_n$), the level of sharpest curvature for the negative predictive value-prevalence curve, such that $\phi_n$ $>$ $\phi_e$. The area between both these thresholds bounds the Matthews Correlation Coefficient (MCC) between $\sqrt{2}/2$ and $\sqrt{2}$. Conversely, the ratio of the maximum possible accuracy to that at any point below the prevalence threshold, $\phi_e$, goes to infinity with decreasing prevalence. Though applications are numerous, the ideas herein discussed may be used in computational complexity theory, artificial intelligence, and medical screening, amongst others. Where computational time is a limiting resource, attaining the prevalence threshold in binary classification systems may be sufficient to yield levels of accuracy comparable to that under maximum prevalence.
\end{abstract} 
\newpage

\section{Background}
\subsection{Classification Systems}
Classification systems categorize elements of a set into different groups on the basis of a classification rule or algorithm \cite{chicco2021matthews}. More specifically, binary classification systems make predictions about a set of elements, placing them into one of two mutually exclusive categories \cite{chicco2020advantages}. In most binary classification problems, one class represents the normal condition (true category) and the other represents the aberrant condition (false category). The classification rules are predetermined and optimized such that the accuracy of the model is maximized \cite{chicco2020advantages}. We define accuracy in this context, henceforth referred to as $\aleph$, as the proportion of correct predictions (both true positives and true negatives) among the total number of cases examined \cite{chicco2020advantages}. The findings of classification systems are summarized in confusion matrices - comparing the predictions made by the model to the true nature of the elements inspected. Depending on the field of study, there may be variation in the terminology used, but a confusion matrix summarizes the performance of a model through the use of statistical metrics \cite{trevethan2017sensitivity} - the most common of which are: the prevalence (relative frequency), sensitivity (recall), specificity (selectivity), positive predictive value/PPV (precision) and negative predictive value (NPV) \cite{trevethan2017sensitivity}. These metrics are defined by the following equations (in bold):
\\
\

\textbf{Figure 1. Confusion Matrix}
\begin{table}[h!]
\centering
\begin{tabular}{llllcl}
                      &                                                                                             &                                                                                                  &                                                                                                  & \multicolumn{1}{l}{}                                                                           &  \\ \cline{3-4}
                      & \multicolumn{1}{l|}{}                                                                       & \multicolumn{1}{c|}{\textbf{\begin{tabular}[c]{@{}c@{}}Observed\\ Positive\end{tabular}}}        & \multicolumn{1}{c|}{\textbf{\begin{tabular}[c]{@{}c@{}}Observed\\ Negative\end{tabular}}}        & \multicolumn{1}{l}{}                                                                           &  \\ \cline{2-5}
\multicolumn{1}{c|}{} & \multicolumn{1}{c|}{\textbf{\begin{tabular}[c]{@{}c@{}}Predicted\\ Positive\end{tabular}}}  & \multicolumn{1}{l|}{True Positive (TP)}                                                          & \multicolumn{1}{l|}{False Negative (FN)}                                                         & \multicolumn{1}{c|}{\textbf{\begin{tabular}[c]{@{}c@{}}PPV\\ TP / TP + FN\end{tabular}}}       &  \\ \cline{2-5}
\multicolumn{1}{l|}{} & \multicolumn{1}{c|}{\textbf{\begin{tabular}[c]{@{}c@{}}Predicted \\ Negative\end{tabular}}} & \multicolumn{1}{l|}{False Positive (FP)}                                                         & \multicolumn{1}{l|}{True Negative (TN)}                                                          & \multicolumn{1}{c|}{\textbf{\begin{tabular}[c]{@{}c@{}}NPV\\ TN / TN + FP\end{tabular}}}       &  \\ \cline{2-5}
                      & \multicolumn{1}{l|}{}                                                                       & \multicolumn{1}{c|}{\textbf{\begin{tabular}[c]{@{}c@{}}Sensitivity\\ TP / TP + FP\end{tabular}}} & \multicolumn{1}{c|}{\textbf{\begin{tabular}[c]{@{}c@{}}Specificity\\ TN / TN + FN\end{tabular}}} & \multicolumn{1}{c|}{\textbf{\begin{tabular}[c]{@{}c@{}}Prevalence\\ TP+FP+FN+TN\end{tabular}}} &  \\ \cline{3-5}
\end{tabular}
\end{table}

\subsection{Properties of Metrics in a Confusion Matrix}
As stated previously, the most frequent metrics used to describe the performance of a binary classification system can be deduced from a confusion matrix  \cite{trevethan2017sensitivity}. All metrics share an important property which is that they can all be defined as a percentage. As such, the sensitivity (recall), specificity (selectivity), positive predictive value (precision), negative predictive value and prevalence (relative frequency) are all defined by a number between 0 and 1  \cite{trevethan2017sensitivity}. Values closer to 1 imply a greater performance of the metric in question and vice-versa \cite{loong2003understanding}. While the precision and negative predictive value are dependent on the prevalence or relative frequency of the elements of a set, the sensitivity/recall and specificity/selectivity are prevalence-independent \cite{manrai2014medicine}.
\newpage
\subsection{Bayes' Theorem and Predictive Values}

Bayes' theorem provides a mathematical framework to explain how existing beliefs change in light of new evidence \cite{efron2013bayes}. From Bayes' theorem, we can derive the positive predictive value/precision, $\rho(\phi)$,  of a binary classification system, defined as the percentage of positive predictions which correctly match the nature of the element assessed, as follows \cite{rouder2018teaching}:

\begin{large}
\begin{equation}
\rho(\phi) = \frac{a\phi}{a\phi+(1-b)(1-\phi)} 
\end{equation}
\end{large} 
\\
\
where $\rho(\phi)$ = PPV/precision, a = sensitivity/recall, b = specificity/selectivity and $\phi$ = prevalence/relative frequency. The PPV, $\rho(\phi)$, is therefore a function of the prevalence/relative frequency, $\phi$. As the prevalence increases, $\rho(\phi)$ also increases and vice-versa.

\subsection{Prevalence Threshold}
We have previously defined the prevalence threshold \cite{balayla2020prevalence} as the prevalence level in the precision-prevalence curve below which binary classification systems start to fail. What does this mean? In technical terms, this point represents the inflection point of maximum curvature in the aforementioned curve below which the the rate of change of a tool's positive predictive value drops at a differential pace relative to the prevalence (or relative frequency) of the elements studied \cite{balayla2021formalism}. Below this point, type I errors -  the mistaken rejection of a true null hypothesis, what is colloquially known as "false positives" - increase. This value, termed $\phi_e$, is a function of a test's sensitivity $a$ and specificity $b$, is defined at the following point on the prevalence axis:
\

\begin{large}
\begin{equation}
\phi_e =\frac{\sqrt{1-b}}{\sqrt{a}+\sqrt{1-b}}
\end{equation}
\end{large}

The corresponding positive predictive value at this prevalence level  is given by plotting the above equation into the positive predictive value equation. In so doing we obtain:

\begin{large}
\begin{equation}
\rho(\phi_e)=\sqrt{\frac{a}{1-b}}\frac{\sqrt{1-b}}{\sqrt{a}+\sqrt{1-b}}=\sqrt{\frac{a}{1-b}}\phi_e
\end{equation}
\end{large}

Interestingly, the above expression leads to the well known formulation for the positive predictive value as  a function of prevalence and the positive likelihood ratio (LR+), defined as the sensitivity $a$ over the compliment of the specificity $b$. Graphically, an example of a precision-prevalence curve with some sensitivity $a$ and specificity $b$, reveals the point of maximum curvature where the radius of curvature $R_c$ is at a minimum:
\\
\

\begin{small}
\begin{center}
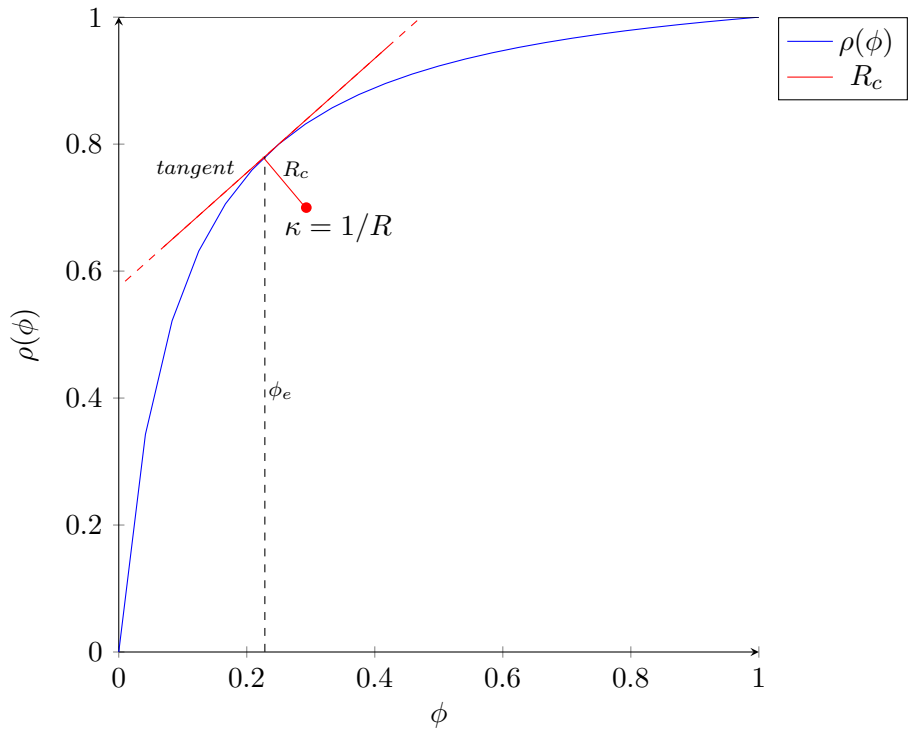
\textbf{Figure 2. Precision-prevalence curve and the prevalence threshold $\phi_e$}
\end{center}
\end{small}
\begin{center}
\begin{tikzpicture}
	\begin{axis}[
    axis lines = left,
    xlabel = $\phi$,
	ylabel = {$\rho(\phi)$},    
     ymin=0, ymax=1,
    legend pos = outer north east,
     ymajorgrids=false,
    grid style=dashed,
    width=10cm,
    height=10cm,
     ]
	\addplot [
	domain= 0:1,
	color= blue,
	]
	{(0.60*x)/((0.60*x+(1-0.95)*(1-x))};

\addplot [	
domain=0.070:0.42, samples=100, color=red] {(0.9*x+0.575)};

\addplot [dashed,	
domain=0.01:0.62, samples=100, color=red] {(0.9*x+0.575)};

\addplot [	
domain=0.225:0.295, samples=100, color=red] {(-1.20*x+1.05005)};

\addplot+ [
domain= 0:1,	
color = black,
mark size = 0pt
 ]
	{1};
	\fill [red] (293,70) circle[radius=2pt];

	\node[above,black] at (343,63) {$\kappa=1/R$};
	\node[above,black] at (120,73) {\scriptsize{$tangent$}};
	\node[above,black] at (277,73) {\scriptsize{$R_c$}};
	\node[above,black] at (253,38) {\scriptsize{$\phi_e$}};
	\addlegendentry{$\rho(\phi)$}
	\addlegendentry{$R_c$}
	\end{axis}
\draw [dashed] (1.92,0) -- (1.92,6.57);
\end{tikzpicture}
\end{center}
\subsection{Accuracy}
Accuracy, $\aleph$, has a number of definitions that depend on the context in which it is used. Generally, accuracy is the condition or quality of being true, correct, or exact; otherwise defined as the closeness of a measured value to a standard or known value \cite{baldi2000assessing}. In essence, it is a measure of correctness when comparing a prediction or observation to an a priori agreed upon 'true' nature of the elements evaluated \cite{baldi2000assessing}. In binary classification systems, accuracy is defined as the proportion of correct predictions - both positive and negative - made by a classification model or computational algorithm \cite{chicco2021benefits}. A value between 0 and 1, often presented as a percentage, the accuracy of a classification model is dependent on several factors, notably: the classification rule or algorithm used, the intrinsic characteristics of the tool used to do the classification, and the relative frequency of the elements being classified \cite{balayla2021formalism}. Several accuracy metrics exist, each with its own advantages in different classification scenarios \cite{kumar2015optimistic}. Herein, we explore several of the most commonly used accuracy metrics, and demonstrate how the prevalence threshold bounds the accuracy of these systems. In other words, when attaining a relative frequency of the elements beyond a particular threshold, the accuracy can only improve so much before attaining values comparable to that under maximum prevalence.
\newpage
\section{Accuracy Metrics}
\subsection{F1 score}
In statistical analysis of binary classification, the F-score or F-measure is a measure of a test's accuracy \cite{chicco2020advantages}. 
As stated previously, accuracy is the proportion of correct predictions ($both$ true positives and true negatives) among the total number of elements inspected. The F1 score is the harmonic mean of the precision and recall \cite{hannun2019cardiologist}. As a metric it does less well in unbalanced groups where the elements being classified appear either too infrequently or too frequently relative to one another. We can determine the generic formula for F1 at any prevalence level through the following harmonic mean:
\\
\
\begin{large}
\begin{equation}
F1 = \frac{2}{\frac{1}{a}+\frac{1}{\rho(\phi)}} = \frac{2}{\frac{1}{a}+\frac{1}{\frac{a\phi}{a\phi+(1-b)(1-\phi)}}} 
\end{equation}
\end{large} 
\\
\
Given equation (3) we can determine the F1-score at the prevalence threshold through the following harmonic mean:
\\
\
\begin{large}
\begin{equation}
F1_{PT} = \frac{2}{\frac{1}{a}+\frac{1}{\sqrt{\frac{a}{1-b}}\frac{\sqrt{1-b}}{\sqrt{a}+\sqrt{1-b}}}} 
\end{equation}
\end{large} 
\\
\
We know that the precision is 1 (100$\%$) when the prevalence or relative frequency is 1 (hence the invariant endpoint at [1,1]). The ensuing F1-score is therefore: 
\\
\
\begin{large}
\begin{equation}
F1_{[1,1]} = \frac{2}{\frac{1}{a}+\frac{1}{1}}
\end{equation}
\end{large}
\\
\
The ratio $F1_{\chi}$ of $F1_{[1,1]}$ over $F1_{PT}$  yields:
\\
\
 \begin{large}
\begin{equation}
F1_{\chi} = \frac{\frac{2}{\frac{1}{a}+\frac{1}{1}}}{\frac{2}{\frac{1}{a}+\frac{1}{\sqrt{\frac{a}{1-b}}\frac{\sqrt{1-b}}{\sqrt{a}+\sqrt{1-b}}}}} = 1 + \frac{\sqrt{a(1-b)}}{1+a}
\end{equation}
\end{large}
\\
\
As the specificity $b$ goes to 1, the ratio of F1 scores goes to 1 – indicating no difference in F1 scores
As the sensitivity $a$ goes to 1 – even if the specificity was set to 0 – the ratio is only 1.5.
As the sensitivity goes to 0 – the ratio is 1. We thus conclude that beyond the PT, the ratio $F1_{\chi}$ of any two $F_{1a}>F_{1b}$ scores is bound between 1 and 1.5, whereas below the PT – this ratio goes to infinity.
\\
\
\begin{large}
\begin{equation}
\boxed{1 \leqslant 1 + \frac{\sqrt{a(1-b)}}{1+a} \leqslant 1.5}
\end{equation}
\end{large}
\newpage
\subsection{F$_\beta$ score}
The F$_\beta$ score is the weighted harmonic mean of precision and recall, reaching its optimal value at 1 and its worst value at 0 \cite{aevermann2021machine}. The $\beta$ parameter determines the weight of recall in the combined score. $\beta$ $<$ 1 lends more weight to precision, while $\beta$ $>$ 1 favours recall ($\beta$ $\rightarrow$ 0 considers only precision, $\beta$ $\rightarrow$ $\infty$ only recall).
\\
\
\begin{large}
\begin{equation}
F{_\beta} = (1+\beta^2) \frac{\rho(\phi)a}{\beta^2\rho(\phi) + a}
\end{equation}
\end{large} 
\\
\
The F$_\beta$ score is the generalized formula for F-score accuracy, of which the F1 score is but a special case when $\beta$ = 1. Therefore, the F$_\beta$ ratio of perfect accuracy to that over the accuracy at the prevalence threshold follows the same steps as for the F1 score and can be simplified to the following:
\\
\
\begin{large}
\begin{equation}
F{_{\beta{_\chi}}} = \frac{(1+\beta^2) \frac{(1)a}{\beta^2(1) + a}}{(1+\beta^2) \frac{{\sqrt{\frac{a}{1-b}}}\frac{\sqrt{1-b}}{\sqrt{a}+\sqrt{1-b}}a}{\beta^2{\sqrt{\frac{a}{1-b}}}\frac{\sqrt{1-b}}{\sqrt{a}+\sqrt{1-b}} + a}} = 1 + \frac{\sqrt{a(1-b)}}{\beta^2+a}
\end{equation}
\end{large} 
\\
\
Though the $\beta$ values can be chosen arbitrarily - the convention is to account for either double or half the precision relative to the recall. As such, the standard convention is to consider values of $\beta$ = 0.5, 1.0 or 2.0. Akin to the F1 score, opting for the different $a$, $b$, $\beta$ values yields a boundary in the domain of $F{_{\beta{_\chi}}}$ such that:
\begin{align*}
\boxed{F{_{\beta{_\chi}}} = 1 + \frac{\sqrt{a(1-b)}}{\beta^2+a} =
\begin{cases} 
      1 \leq F{_{\beta{_\chi}}} \leq 1.8  & \text{if } \beta = 0.5 \\
      1 \leq F{_{\beta{_\chi}}} \leq 1.5  & \text{if } \beta = 1 \\
     1 \leq F{_{\beta{_\chi}}} \leq 1.2 & \text{if } \beta = 2 
\end{cases}}
\end{align*}
\\
\

We observe here too that the $F{_{\beta{_\chi}}}$ is bound between 1 and 1.8 depending on the value of $\beta$ chosen a priori.
\\
\

\subsection{Fowlkes–Mallows Index (FM)}
The Fowlkes–Mallows (FM) index \cite{fowlkes1983method} is an external evaluation method that is used to determine the similarity between two clusterings or sets, and is also a metric to evaluate accuracy through confusion matrices. Like the F1 score, the FM index is a function of the precision and recall, except that instead of the harmonic mean, the FM index is the geometric mean of precision and recall. The minimum possible value of the Fowlkes–Mallows index is 0, which corresponds to the worst binary classification possible, where all the elements have been misclassified. Likewise, the maximum possible value of the Fowlkes–Mallows index is 1, which corresponds to the best binary classification possible, where all the elements have been perfectly classified \cite{fowlkes1983method}. We can determine the generic formula for FM at any prevalence level through the following equation:
\\
\
\begin{large}
\begin{equation}
FM = \sqrt{a\cdot\rho(\phi)}
\end{equation}
\end{large}
\\
\
We repeat the same exercise as for the F1 and calculate the FM at the PT and the FM at the endpoint [1,1] to determine the impact of the PT on the FM evaluation:
\\
\
\begin{large}
\begin{equation}
FM_{PT} = \sqrt{a\left(\sqrt{\frac{a}{1-b}}\frac{\sqrt{1-b}}{\sqrt{a}+\sqrt{1-b}}\right)}
\end{equation}
\end{large}
\\
\
\begin{large}
\begin{equation}
FM_{[1,1]} = \sqrt{a\cdot(1)}
\end{equation}
\end{large}
\\
\
The ratio $FM_{\chi}$ of $FM_{[1,1]}$ over $FM_{PT}$  yields:
\\
\
 \begin{large}
\begin{equation}
FM_{\chi}=\frac{\sqrt{a(1)}}{\sqrt{a\left(\sqrt{\frac{a}{1-b}}\frac{\sqrt{1-b}}{\sqrt{a}+\sqrt{1-b}}\right)}} = \sqrt{1+\sqrt{\frac{1-b}{a}}}
\end{equation}
\end{large}
\\
\
We thus conclude, based on equation (14), that beyond the prevalence threshold the FM index is bound between 1 and $\sqrt{2}$. 
\\
\
\begin{large}
\begin{equation}
\boxed{1\leqslant\sqrt{1+\sqrt{\frac{1-b}{a}}}\leqslant\sqrt{2}}
\end{equation}
\end{large}
\\
\
Here too, below the PT – this ratio goes to infinity. We thus make a generalized statement about the performance of binary classification systems below the prevalence threshold relative to the perfect accuracy $\aleph_1$, as follows:
\\
\
\begin{large}
\begin{equation}
\lim_{\aleph_x \to 0} \chi = \frac{\aleph_1}{\aleph_x}=\infty
\end{equation}
\end{large}
\\
\
where $\aleph_x$ $<$ $\aleph_e$ $<$ $\aleph_1$. We can better appreciate equation 16 graphically, taking an example classification system with sensitivity 0.90 and specificity 0.95. Note the dotted line in black, which represents the positive prevalence threshold, below which $\chi = \aleph_1/\aleph_e$ goes to infinity. Note likewise how beyond the prevalence threshold the ratio converges to 1 for all metrics in question.
\newpage
\begin{center}
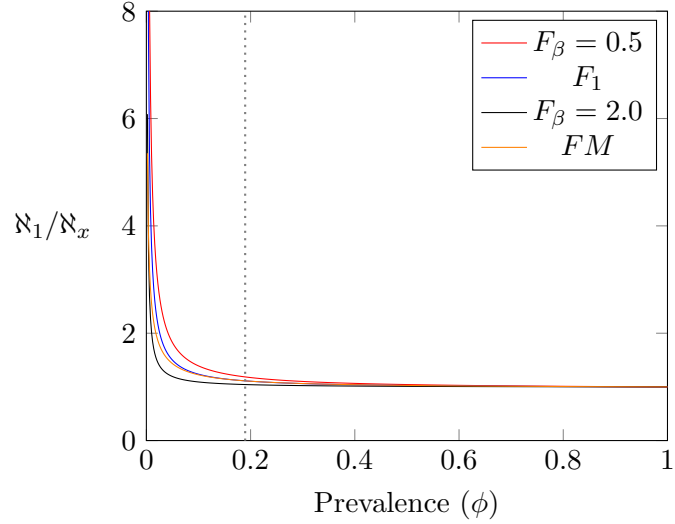
\textbf{Figure 3. Graphical representation of $\aleph_1/\aleph_x$ as a function of prevalence $\phi$ for different accuracy metrics}
\\
\

\begin{tikzpicture}
\begin{axis}
[
    xlabel={Prevalence ($\phi$)}, 
    ylabel={$\aleph_1/\aleph_x$},  
    xmin=0, 
    xmax=1, 
    ymin=0, 
    ymax=8, 
    y label style={rotate=-90}
]
\addplot[red, no marks, domain=0:1, samples=500] {(0.78260869565)/((0.81*x)/((0.85*x+0.05)*((0.225*x)/(0.85*x +0.05)+0.9)))};
\addlegendentry{$F_{\beta}=0.5$}

\addplot[blue, no marks, domain=0:1, samples=500] {(0.9*x+0.95*x-0.95+1)/(x*(0.90+1))};
\addlegendentry{$F_1$}

\addplot[black, no marks, domain=0:1, samples=500] {(0.18367)/((0.81*x)/((0.85*x+0.05)*((3.6*x)/(0.85*x +0.05)+0.9)))};
\addlegendentry{$F_{\beta}=2.0$}

\addplot[orange, no marks, no marks, domain=0:1, samples=500] {((0.85*x +0.05)^(0.5))/(0.9*x)^(0.5)};
\addlegendentry{$FM$}

\end{axis}
\draw [thick,dotted,gray] (1.3,0) -- (1.3,5.70);
\end{tikzpicture}
\end{center}

\subsection{Matthews Correlation Coefficient (MCC)}
The Matthews Correlation Coefficient, or MCC for short, is a more comprehensive accuracy metric of binary classifications systems \cite{chicco2021matthews}. The MCC takes into account all entries in a confusion matrix and as such, is better poised to address asymmetries or imbalances in the model data. Related to the Chi-Square ($\chi^2$) statistic for 2x2 contingency tables, the MCC is also known the $\Phi$ coefficient and its interpretation is similar to that of the Pearson correlation coefficient \cite{de2018method}, where values range from -1 (total disagreement between predictions and observations) to 0 (no better than random predictions) and +1 (perfect concordance between prediction and observation).
\\
\
\begin{large}
\begin{equation}
|\Phi| = \sqrt{\frac{\chi^2}{n}}
\end{equation}
\end{large}
\\
\
The MCC can be defined as:
\\
\
\begin{equation}
MCC = \sqrt{\rho\cdot a \cdot b \cdot \sigma} - \sqrt{(1-\rho)\cdot(1-a)\cdot(1-b)\cdot(1-\sigma)}
\end{equation}
\\
\
where $\rho$ = precision, a = recall, b = specificity, $\sigma$ = NPV.
\\
\

We would first need to define the equation for the negative predictive value, $\sigma(\phi)$, as a function of prevalence - which can also be derived from Bayes' Theorem:
\\
\
\begin{large}
\begin{equation}
\sigma(\phi) = \frac{b(1-x)}{b(1-x)+(1-a)x}
\end{equation}
\end{large}
\\
\
Given the Bayesian properties of the predictive value-prevalence curves \cite{hall1967clinical} - the methods we have used thus far to compare accuracy metrics at the prevalence threshold vs. at the maximum prevalence will fail to produce meaningful results since at the maximum prevalence, the negative predictive value, $\sigma(\phi)$ is 0  (see red curve in Figure 4.)\cite{balayla2020prevalence}. The ensuing product would thus be 0, rendering no useful information about the relative performance of this metric. To overcome this hurdle, we can define the $negative$ prevalence threshold - the prevalence level beyond which the NPV curve drops most significantly, and compare the performance of the MCC at both thresholds. The two different prevalence thresholds can perhaps be best visualized in the following graphic:

\begin{center}
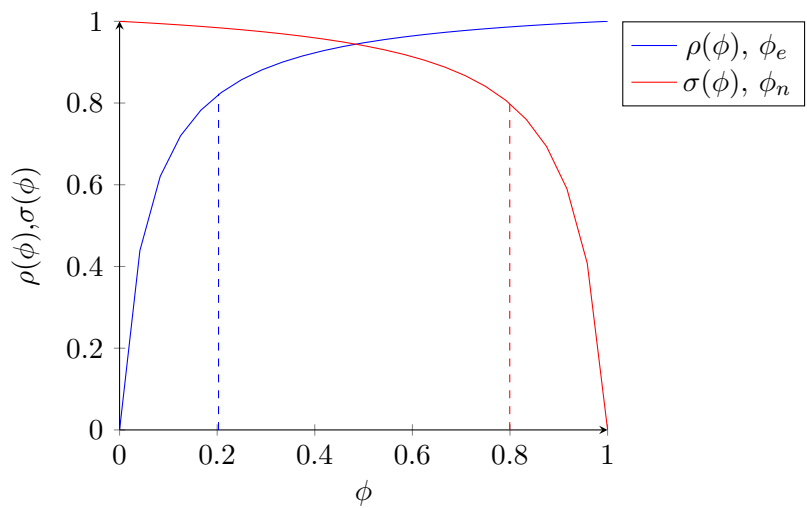
\textbf{Figure 4. Positive ($\phi_e$) and negative ($\phi_n$) prevalence thresholds}
\end{center}
\begin{center}
\begin{tikzpicture}
	\begin{axis}[
    axis lines = left,
    xlabel = $\phi$,
	ylabel = {$\rho(\phi)$,$\sigma(\phi)$},    
     ymin=0, ymax=1,
    legend pos = outer north east,
     ymajorgrids=false,
     xmajorgrids=false,
    width=8cm,
    height=7cm,
     ]
	\addplot [
	domain= 0:1,
	color= blue,
	]
	{(0.90*x)/((0.90*x+(1-0.95)*(1-x))};

\addplot [	
	domain= 0:1,
	color= red,
	]
	{0.95*(1-x)/((1-0.94)*x+0.95*(1-x))};	
	\addlegendentry{$\rho(\phi)$, $\phi_e$}
	\addlegendentry{$\sigma(\phi)$, $\phi_n$}
	\end{axis}
	\draw [dashed,blue] (1.3,0) -- (1.3,4.46);
	\draw [dashed,red] (5.13,0) -- (5.13,4.4);
\end{tikzpicture}
\end{center}

The derivation of the $negative$ prevalence threshold is beyond the scope of this paper but can be observed in the red dashed line (Figure 4), and is defined as level of maximum curvature of the NPV-prevalence curve:

\begin{large}
\begin{equation}
R = \frac{1}{\kappa} \Rightarrow \kappa = \frac{|\frac{d^2\rho}{d\phi^2}|}{[1 + (\frac{d\rho}{d\phi})^2]^\frac{3}{2}} \rightarrow \frac{d\kappa}{d\phi}=0 \hookrightarrow\lbrace\phi_n,\rho_n\rbrace 
\end{equation}
\end{large}
\\
\
The $negative$ prevalence threshold is thus:
\begin{large}
\begin{equation}
\phi_n =\frac{\sqrt{b}}{\sqrt{1-a}+\sqrt{b}}
\end{equation}
\end{large}
\newpage
Note that for all values of recall and specificity, where $\varepsilon$ = a + b $>$ 1, $\phi_n>\phi_e$. Thus, using this new prevalence level ($\phi_n$) instead of 1, we can now calculate the ratio of the MCC at $\phi_n$ vs. $\phi_e$ as follows:
\\
\
\begin{small}
\begin{equation}
\begin{aligned}
MCC_{\phi_e} = {} &\sqrt{\sqrt{\frac{a}{1-b}}\frac{\sqrt{1-b}}{\sqrt{a}+\sqrt{1-b}}\cdot a \cdot b \cdot \frac{b(1-\frac{\sqrt{1-b}}{\sqrt{a}+\sqrt{1-b}})}{b(1-\frac{\sqrt{1-b}}{\sqrt{a}+\sqrt{1-b}})+(1-a)\frac{\sqrt{1-b}}{\sqrt{a}+\sqrt{1-b}}}}... \\ &- \sqrt{(1-\sqrt{\frac{a}{1-b}}\frac{\sqrt{1-b}}{\sqrt{a}+\sqrt{1-b}})\cdot(1-a)\cdot(1-b)\cdot(1-\frac{b(1-\frac{\sqrt{1-b}}{\sqrt{a}+\sqrt{1-b}})}{b(1-\frac{\sqrt{1-b}}{\sqrt{a}+\sqrt{1-b}})+...}}...)
\end{aligned}
\end{equation}
\end{small}
\\
\
We likewise adopt a similar approach and input the appropriate variables to obtain the full $MCC_{\phi_n}$ equation. We can calculate the ratio of the MCC at the $\phi_n$ over the $\phi_e$ to obtain (full equation in the addendum):

\begin{equation}
\begin{aligned}
MCC_{\chi} = \frac{MCC_{\phi_n}}{MCC_{\phi_e}}
\end{aligned}
\end{equation}
\\
\
Given the difference of square roots in both numerator and denominator, we could likewise simplify the equations as follows:
\begin{equation}
MCC_{\chi}=\frac{\sqrt{A}-\sqrt{B}}{\sqrt{C}-\sqrt{D}}
\end{equation}
where
\begin{align*}
A &= a\cdot b \cdot c \cdot d \\
B &= (1-a)(1-b)(1-c)(1-d) \\
C &= a\cdot b\cdot e\cdot f  \\
D &= (1-a)(1-b)(1-e)(1-f) \\
\shortintertext{and}
a &= {sensitivity}\\
b &= {specificity}\\
u &= \sqrt{b}\big/\bigl(\sqrt{1-a}+\sqrt{b}\,\bigr)\\
v &= \sqrt{1-b}\big/\bigl(\sqrt{a}+\sqrt{1-b}\,\bigr)\\
c &= au\big/\bigl(au+(1-b)(1-u)\bigr) \\
d &= \b(1-u)\big/\bigl(b(1-u)+(1-a)u\bigr)\\
e &= b(1-v)\big/\bigl(b(1-v)+(1-a)v\bigr)\\
f &= \sqrt{a/(1-b)}v\\,
\end{align*}
Given the above ratio, the $MCC_{\chi}$, the ratio of MCC at the negative prevalence threshold over the positive prevalence threshold is bound by:
\\
\
\begin{large}
\begin{equation}
\boxed{\frac{\sqrt{2}}{2} \leqslant MCC_{\chi} \leqslant \sqrt{2}}
\end{equation}
\end{large}
\\
\
Unlike the other accuracy metrics, note the lower bound of $MCC_{\chi}$ can be $<$ 1. This is due to the fact that with increasing prevalence, the positive predictive value increases but the negative predictive value decreases. Depending on the values of sensitivity and specificity in a given context, the $MCC_{\chi}$ may decrease with increasing prevalence but is nevertheless likewise bound as are the other metrics.
\\
\
\section{Applications}
Aside from an enhancement in the understanding of the theoretical nature of accuarcy metrics, the applications of the concepts herein described are multiple. For example, understanding the value of the prevalence threshold as it pertains to accuracy may render it a more attainable objective when attempting to reach acceptable levels of accuracy in a classification system. Indeed, computational complexity theory focuses on classifying computational problems according to their resource usage, and relating these classes to each other \cite{hartmanis1965computational}. A problem is regarded as inherently difficult if its solution requires significant resources, whatever the algorithm used. The theory formalizes this intuition, by introducing a mathematical model of computation to study these problems and quantifying their computational complexity, i.e., the amount of resources needed to solve them, such as time and storage \cite{hartmanis1965computational}. As we saw throughout this manuscript, we can trade computational time or resources for minimal loss in accuracy by using the prevalence threshold as an endpoint in binary classification problems since by definition $\phi_e$ $<$ 1 $\rightarrow$ $\aleph_e$ $<$ $\aleph_1$. Likewise, in artificial intelligence, classification models are often binary - depending on the scenario, we may adjust the prevalence, which is not always possible, or instead, we may adopt the prevalence threshold as an endpoint that may serve as a means to train AI models faster, and to verify classification outcomes without needing to enhance the prevalence significantly. Finally, binary classification systems lie at the core of modern medicine and public health, whereby public health screening programs categorize individuals according to binary disease status (sick vs. $not$ sick). When modelling disease to adopt public health measures, using the prevalence threshold and its bounds on the accuracy of binary classification systems may provide critical information about the management of disease in a population where the prevalence is unknown - since as we have seen in this manuscript - the prevalence threshold, which can be calculated using only a system's recall and specificity, yields accuracy information akin to that under maximum prevalence.
\section{Conclusion}
In this manuscript, we have shown that the relative ratio of maximum accuracy of binary classification systems to levels below the prevalence threshold goes to infinity, whereas beyond this limit, the ratio is bound. Attaining the prevalence threshold in binary classification systems may be sufficient to yield levels of accuracy comparable to that under maximum prevalence.
\newpage
\section{Addendum}
\
In its long form, the $MCC_{\chi}$ equation can be defined as follows:
\begin{tiny}
\begin{equation}
MCC_{\chi}=
\begin{split}
&\left(\sqrt{\frac{a\frac{\sqrt{b}}{\sqrt{1-a}+\sqrt{b}}}
{a\frac{\sqrt{b}}{\sqrt{1-a}+\sqrt{b}}+\left(1-b\right)
\left(1-\frac{\sqrt{b}}{\sqrt{1-a}+\sqrt{b}}\right)}ab
\frac{b\left(1-\frac{\sqrt{b}}{\sqrt{1-a}+\sqrt{b}}\right)}
{b\left(1-\frac{\sqrt{b}}{\sqrt{1-a}+\sqrt{b}}\right)+\left(1-a\right)
\frac{\sqrt{b}}{\sqrt{1-a}+\sqrt{b}}}}\right.\\
&\left.-\sqrt{\left(1-\frac{a\frac{\sqrt{b}}{\sqrt{1-a}+\sqrt{b}}}
{a\frac{\sqrt{b}}{\sqrt{1-a}+\sqrt{b}}+\left(1-b\right)
\left(1-\frac{\sqrt{b}}{\sqrt{1-a}+\sqrt{b}}\right)}\right)\left(1-a\right)
\left(1-b\right)\left(1-\frac{b\left(1-\frac{\sqrt{b}}{\sqrt{1-a}+\sqrt{b}}\right)}
{b\left(1-\frac{\sqrt{b}}{\sqrt{1-a}+\sqrt{b}}\right)+\left(1-a\right)
\frac{\sqrt{b}}{\sqrt{1-a}+\sqrt{b}}}\right)}\right)\\
\Bigg/&\left(\sqrt{\sqrt{\frac{a}{1-b}}\frac{\sqrt{1-b}}{\sqrt{a}+\sqrt{1-b}}ab
\frac{b\left(1-\frac{\sqrt{1-b}}{\sqrt{a}+\sqrt{1-b}}\right)}
{b\left(1-\frac{\sqrt{1-b}}{\sqrt{a}+\sqrt{1-b}}\right)+\left(1-a\right)
\frac{\sqrt{1-b}}{\sqrt{a}+\sqrt{1-b}}}}\right.\\
&\left.-\sqrt{\left(1-\sqrt{\frac{a}{1-b}}\frac{\sqrt{1-b}}{\sqrt{a}+\sqrt{1-b}}\right)
\left(1-a\right)\left(1-b\right)
\left(1-\frac{b\left(1-\frac{\sqrt{1-b}}{\sqrt{a}+\sqrt{1-b}}\right)}
{b\left(1-\frac{\sqrt{1-b}}{\sqrt{a}+\sqrt{1-b}}\right)+\left(1-a\right)
\frac{\sqrt{1-b}}{\sqrt{a}+\sqrt{1-b}}}\right)}\right)
\end{split}
\end{equation}
\end{tiny}
\\
\
Replacing the threshold values by $\phi_n$ and $\phi_e$ we obtain:
\
\begin{small}
\begin{equation}
MCC_{\chi}= \frac{\sqrt{\frac{a\phi_n}{a\phi_n + (1-b)(1-\phi_n)} ab \frac{b(1-\phi_n)}{b(1-\phi_n)+(1-a)\phi_n}} - \sqrt{\left(1 - \frac{a\phi_n}{a\phi_n + (1-b)(1-\phi_n)}\right) (1-a)(1-b)\left(1-\frac{b(1-\phi_n)}{b(1-\phi_n) + (1-a)\phi_n}\right)}%
}{\sqrt{\sqrt{\frac{a}{1-b}} \phi_e ab \frac{b(1-\phi_e )}{b(1-\phi_e ) + (1-a)\phi_e }}-\sqrt{\left(1-\sqrt{\frac{a}{1-b}}\phi_e \right) (1-a)(1-b)\left(1 - \frac{b(1-\phi_e )}{b(1-\phi_e ) + (1-a)\phi_e } \right)}%
}
\end{equation}
\end{small}
where
\[
\phi_n  = \frac{\sqrt{b}}{\sqrt{1-a}+\sqrt{b}}
\qquad\text{and}\qquad
\phi_e  = \frac{\sqrt{1-b}}{\sqrt{a}+\sqrt{1-b}}
\]
\\
\
\newpage

\bibliographystyle{unsrt}
\bibliography{References.bib}

\begin{thebibliography}{10}

\bibitem{chicco2021matthews}
Davide Chicco, Niklas T{\"o}tsch, and Giuseppe Jurman.
\newblock The matthews correlation coefficient (mcc) is more reliable than
  balanced accuracy, bookmaker informedness, and markedness in two-class
  confusion matrix evaluation.
\newblock {\em BioData mining}, 14(1):1--22, 2021.

\bibitem{chicco2020advantages}
Davide Chicco and Giuseppe Jurman.
\newblock The advantages of the matthews correlation coefficient (mcc) over f1
  score and accuracy in binary classification evaluation.
\newblock {\em BMC genomics}, 21(1):1--13, 2020.

\bibitem{trevethan2017sensitivity}
Robert Trevethan.
\newblock Sensitivity, specificity, and predictive values: foundations,
  pliabilities, and pitfalls in research and practice.
\newblock {\em Frontiers in public health}, 5:307, 2017.

\bibitem{loong2003understanding}
Tze-Wey Loong.
\newblock Understanding sensitivity and specificity with the right side of the
  brain.
\newblock {\em Bmj}, 327(7417):716--719, 2003.

\bibitem{manrai2014medicine}
Arjun~K Manrai, Gaurav Bhatia, Judith Strymish, Isaac~S Kohane, and Sachin~H
  Jain.
\newblock Medicine’s uncomfortable relationship with math: calculating
  positive predictive value.
\newblock {\em JAMA internal medicine}, 174(6):991--993, 2014.

\bibitem{efron2013bayes}
Bradley Efron.
\newblock Bayes' theorem in the 21st century.
\newblock {\em Science}, 340(6137):1177--1178, 2013.

\bibitem{rouder2018teaching}
Jeffrey~N Rouder and Richard~D Morey.
\newblock Teaching bayes’ theorem: Strength of evidence as predictive
  accuracy.
\newblock {\em The American Statistician}, 2018.

\bibitem{balayla2020prevalence}
Jacques Balayla.
\newblock Prevalence threshold ($\phi$ e) and the geometry of screening curves.
\newblock {\em Plos one}, 15(10):e0240215, 2020.

\bibitem{balayla2021formalism}
Jacques Balayla.
\newblock On the formalism of the screening paradox.
\newblock {\em Plos one}, 16(9):e0256645, 2021.

\bibitem{baldi2000assessing}
Pierre Baldi, S{\o}ren Brunak, Yves Chauvin, Claus~AF Andersen, and Henrik
  Nielsen.
\newblock Assessing the accuracy of prediction algorithms for classification:
  an overview.
\newblock {\em Bioinformatics}, 16(5):412--424, 2000.

\bibitem{chicco2021benefits}
Davide Chicco, Valery Starovoitov, and Giuseppe Jurman.
\newblock The benefits of the matthews correlation coefficient (mcc) over the
  diagnostic odds ratio (dor) in binary classification assessment.
\newblock {\em IEEE Access}, 9:47112--47124, 2021.

\bibitem{kumar2015optimistic}
S~Senthil Kumar and H~Hannah Inbarani.
\newblock Optimistic multi-granulation rough set based classification for
  medical diagnosis.
\newblock {\em Procedia Computer Science}, 47:374--382, 2015.

\bibitem{hannun2019cardiologist}
Awni~Y Hannun, Pranav Rajpurkar, Masoumeh Haghpanahi, Geoffrey~H Tison, Codie
  Bourn, Mintu~P Turakhia, and Andrew~Y Ng.
\newblock Cardiologist-level arrhythmia detection and classification in
  ambulatory electrocardiograms using a deep neural network.
\newblock {\em Nature medicine}, 25(1):65--69, 2019.

\bibitem{aevermann2021machine}
Brian Aevermann, Yun Zhang, Mark Novotny, Mohamed Keshk, Trygve Bakken, Jeremy
  Miller, Rebecca Hodge, Boudewijn Lelieveldt, Ed~Lein, and Richard~H
  Scheuermann.
\newblock A machine learning method for the discovery of minimum marker gene
  combinations for cell type identification from single-cell rna sequencing.
\newblock {\em Genome research}, 31(10):1767--1780, 2021.

\bibitem{fowlkes1983method}
Edward~B Fowlkes and Colin~L Mallows.
\newblock A method for comparing two hierarchical clusterings.
\newblock {\em Journal of the American statistical association},
  78(383):553--569, 1983.

\bibitem{de2018method}
Paulo~Vitor de~Campos~Souza, Vanessa~Souza Araujo, Augusto~Junio Guimaraes,
  Vincius Jonathan~Silva Araujo, and Thiago~Silva Rezende.
\newblock Method of pruning the hidden layer of the extreme learning machine
  based on correlation coefficient.
\newblock In {\em 2018 IEEE Latin American conference on computational
  intelligence (LA-CCI)}, pages 1--6. IEEE, 2018.

\bibitem{hall1967clinical}
GH~Hall.
\newblock The clinical application of bayes'theorem.
\newblock {\em The Lancet}, 290(7515):555--557, 1967.

\bibitem{hartmanis1965computational}
Juris Hartmanis and Richard~E Stearns.
\newblock On the computational complexity of algorithms.
\newblock {\em Transactions of the American Mathematical Society},
  117:285--306, 1965.

\end{thebibliography}
\end{document}